\documentclass{article}

 \usepackage[preprint]{neurips_2025}

\usepackage[utf8]{inputenc} 
\usepackage[T1]{fontenc}    
\usepackage{hyperref}       
\usepackage{url}            
\usepackage{booktabs}       
\usepackage{amsfonts}       
\usepackage{nicefrac}       
\usepackage{microtype}      
\usepackage[table]{xcolor}
\usepackage{graphicx}
\usepackage{amsmath}
\usepackage{makecell}
\usepackage{tcolorbox}
\usepackage{colortbl}
\usepackage{wrapfig}
\usepackage{pifont}
\usepackage{bbding}
\usepackage{multirow}
\usepackage{enumitem}
\usepackage{ulem}
\usepackage{tabularx}
\usepackage{array}
\newcolumntype{Y}{>{\centering\arraybackslash}X}  
\newcolumntype{Z}{>{\raggedleft\arraybackslash}X}  
\usepackage{textcomp}

\title{ML-Master: Towards AI-for-AI via Integration of Exploration and Reasoning}

\author{%
    Zexi Liu\thanks{Equal contribution. Order randomized.}, \ 
    Yuzhu Cai\footnotemark[1], \ 
    Xinyu Zhu\footnotemark[1],  \ 
    Yujie Zheng\footnotemark[1],  \ 
    Runkun Chen\footnotemark[1],  \\ 
    \textbf{Ying Wen,} \
    \textbf{Yanfeng Wang,} \
    \textbf{Weinan E,} \
    \textbf{Siheng Chen}
\\
    School of Artificial Intelligence, Shanghai Jiao Tong University
}

\begin{document}

\maketitle
\renewcommand{\thefootnote}{\fnsymbol{footnote}}
\addtocounter{footnote}{1}
\footnotetext{Project is available here: \url{https://sjtu-sai-agents.github.io/ML-Master}.}

\begin{figure}[htbp]
    \centering
    \includegraphics[width=0.9\textwidth]{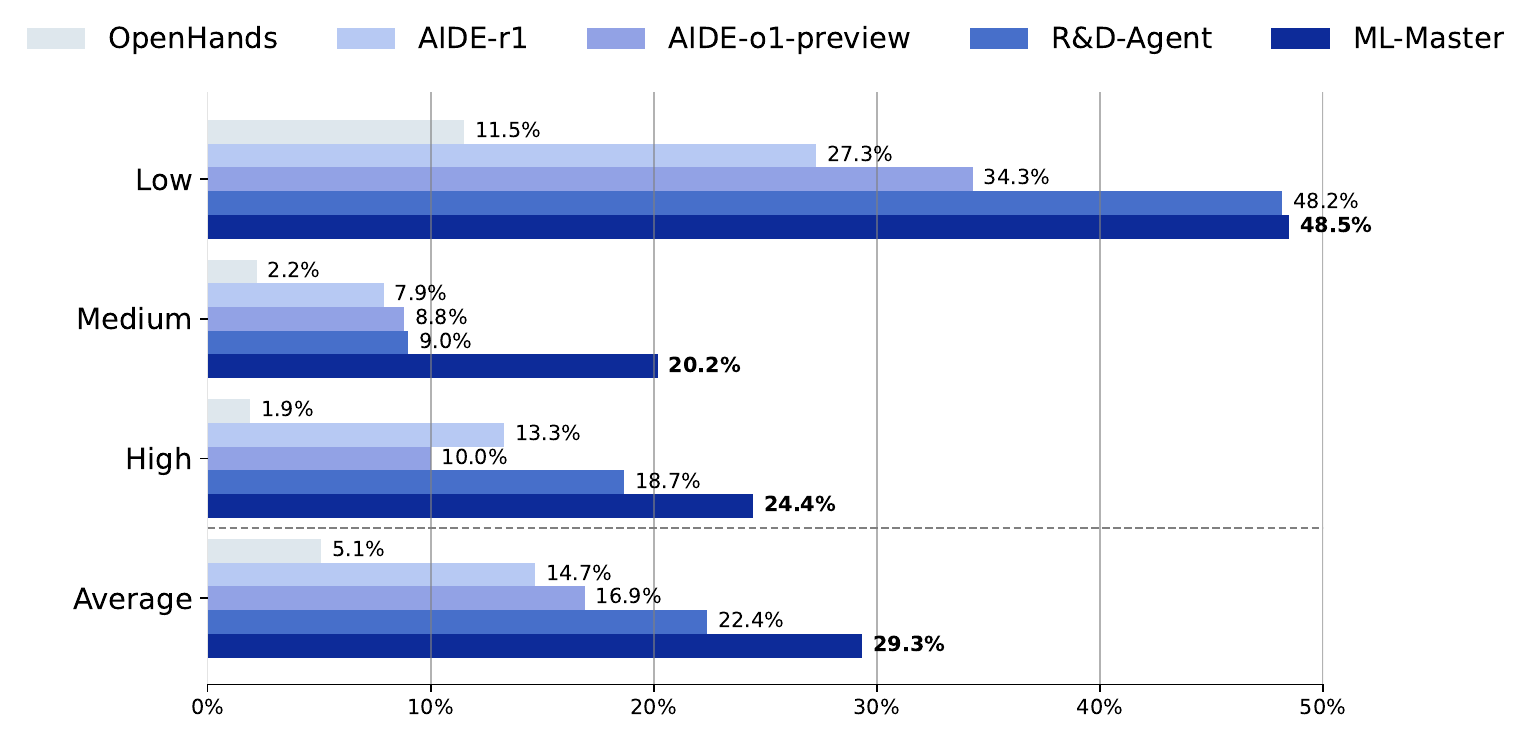}
    \caption{Performance of OpenHands~\cite{openhands}, AIDE~\cite{aide}, R\&D-Agent~\cite{rdagent} and ML-Master on MLE-Bench~\cite{mlebench}.}
    \label{fig:pre_abstract}
\end{figure}

\begin{abstract}
As AI capabilities advance toward and potentially beyond human-level performance, a natural transition emerges where AI-driven development becomes more efficient than human-centric approaches. A promising pathway toward this transition lies in AI-for-AI (AI4AI), which leverages AI techniques to automate and optimize the design, training, and deployment of AI systems themselves. While LLM-based agents have shown the potential to realize AI4AI, they are often unable to fully leverage the experience accumulated by agents during the exploration of solutions in the reasoning process, leading to inefficiencies and suboptimal performance. 
To address this limitation, we propose ML-Master, a novel AI4AI agent that seamlessly integrates exploration and reasoning by employing a selectively scoped memory mechanism. 
This approach allows ML-Master to efficiently combine diverse insights from parallel solution trajectories with analytical reasoning, guiding further exploration without overwhelming the agent with excessive context. 
We evaluate ML-Master on the MLE-Bench, where it achieves a 29.3\% average medal rate, significantly surpassing existing methods, particularly in medium-complexity tasks, 
while accomplishing this superior performance within a strict 12-hour time constraint—half the 24-hour limit used by previous baselines.
These results demonstrate ML-Master's potential as a powerful tool for advancing AI4AI.
\end{abstract}

\section{Introduction}
\vspace{-0.8em}
Artificial Intelligence (AI) has profoundly reshaped human civilization, driving transformative advancements across diverse areas~\cite{szczepanski2019economic,wolff2020economic,handa2025economic,abulibdeh2024navigating}. 
As AI advances toward surpassing human-level intelligence, it is crucial to advocate for the emergence of AI-for-AI (AI4AI), which leverages AI techniques to automate and optimize the design, training, and deployment of AI systems themselves~\cite{zoph2017neuralarchitecturesearch,sumers2024cognitivearchitectureslanguageagents}. AI-for-AI can be understood as a progressive paradigm unfolding in three stages: human-led human-AI collaboration, AI-led human-AI collaboration, and fully autonomous AI systems. We are likely in the midst of transitioning from the first to the second stage. The ultimate form of AI-for-AI envisions fully autonomous systems capable of end-to-end AI research and development—from hypothesis generation and experimental design to algorithmic discovery and validation.  One possible vision for the development of AI4AI can be drawn from the progression seen in systems like AlphaGo~\cite{alphago} and AlphaZero~\cite{alpha_zero}, where the journey began with human-guided training to improve the machine's Go-playing skills, eventually leading to machines surpassing human-level performance and further enhancing their skills through self-play.

Realizing the AI4AI vision begins with understanding how human experts design AI systems. Inspired by how expert AI practitioners work, we observe that developing effective AI solutions is inherently an iterative and exploratory process. 
AI practitioners naturally integrate exploration and reasoning into a cohesive cognitive methodology. 
Specifically, exploration entails actively seeking new insights through various experimentation and discovery~\cite{aide}, 
while reasoning involves carefully analyzing existing knowledge and reflecting upon past experiences~\cite{deepseekr1,openai2024o1}. 
Neither alone is sufficient:  
exploration without reasoning can lead to inefficiency and aimless trial-and-error, while reasoning without exploration risks stagnation.
Instead, effective problem-solving emerges from a harmonious interplay between exploration and reasoning, where new insights gained through exploration continuously enrich and refine subsequent reasoning processes. 
This iterative cycle of purposeful exploration and thoughtful reasoning forms the foundation of continuous improvement and innovation in human-driven AI development, motivating the need for AI4AI frameworks that similarly integrate these complementary cognitive strategies. 

Although recent breakthroughs in Large Language Models (LLMs)~\cite{deepseekr1,openai2024o1,claude4} and autonomous agents~\cite{openhands,zhang2023automl,zheng2025deepresearcherscalingdeepresearch,agentlaboratory} have provided evidence supporting the feasibility of AI4AI, most studies still encounter significant challenges. 
Previous works~\cite{ai_scientist,sela,mlagentbench,dophin,paper2code} like AI Scientist~\cite{ai_scientist}, SELA~\cite{sela}, and Dolphin~\cite{dophin} primarily emphasize exploration strategies without sufficiently leveraging the analytical reasoning capabilities of advanced reasoning models, thus missing valuable insights and limiting their adaptability in complex scenarios. 
Conversely, works such as AIDE~\cite{aide} and Agent Laboratory~\cite{agentlaboratory} attempt to utilize the reasoning capabilities, but their exploration strategies are often inefficient or insufficiently comprehensive, leading to hallucinations, unreliable outputs, and suboptimal performance. 
Overall, existing AI4AI methods struggle to effectively integrate exploration and reasoning, 
primarily because exploration processes often fail to sufficiently distill past experiences to generate promising solutions. 
Additionally, reasoning models find it challenging to effectively utilize the extensive and unstructured experiences accumulated during exploration, as overly long contexts can overwhelm the reasoning process, leading to hallucinations and unreliable outputs.

To bridge the gap between exploration and reasoning, 
we introduce \textbf{ML-Master}, a novel AI4AI agent inspired by the unified cognitive strategies of expert AI developers. 
Unlike exsiting AI4AI methods as summarized in Table~\ref{tab:comparison}, 
ML-Master \uline{integrates exploration and reasoning into a cohesive iterative methodology}
by employing an adaptive memory mechanism that selectively captures and summarizes insights from exploration history. This design ensures each component mutually reinforces the other without compromising either. 
Specifically, ML-Master simultaneously leverages the analytical learning capabilities of reasoning models and a comprehensive, efficient exploration strategy, forming a virtuous cycle of continuous improvement. 
Within this unified cognition, ML-Master comprises two complementary and mutually supportive modules:
(1) \textbf{Balanced multi-trajectory exploration} empowers ML-Master to explore multiple solution trajectories step by step in parallel while maintaining optimal balance between exploitation of promising paths and exploration of under-investigated alternatives. 
By dynamically prioritizing trajectories based on their potential value and exploration history, this module enables ML-Master to actively generate diverse experiences and insights without over-committing to any single direction.   
These exploratory outcomes form a memory consisted of concrete execution feedback and new knowledge to enrich the reasoning process, enabling more informed and accurate analytical reasoning in subsequent iterations. 
(2) \textbf{Steerable reasoning} enhances the reasoning capabilities of an advanced reasoning model (Deepseek-R1~\cite{deepseekr1}) by explicitly embedding the adaptive memory into the reasoning process. 
Insights and execution feedback from exploration trajectories are selectively incorporated into the reasoning process, 
enabling the model to learn from past experiences while avoiding redundant reasoning paths.
This integration ensures precise, reliable, and controlled analytical capabilities, significantly reducing hallucinations and erroneous interpretations commonly observed in LLM-based agents. 
The insights derived from this reasoning process directly inform and guide exploration steps.
By organically combining balanced multi-trajectory exploration and steerable reasoning within a unified framework, ML-Master achieves robust, efficient, and high-performing AI4AI.

\begin{figure}[!t]
    \centering
    \includegraphics[width=0.9\textwidth]{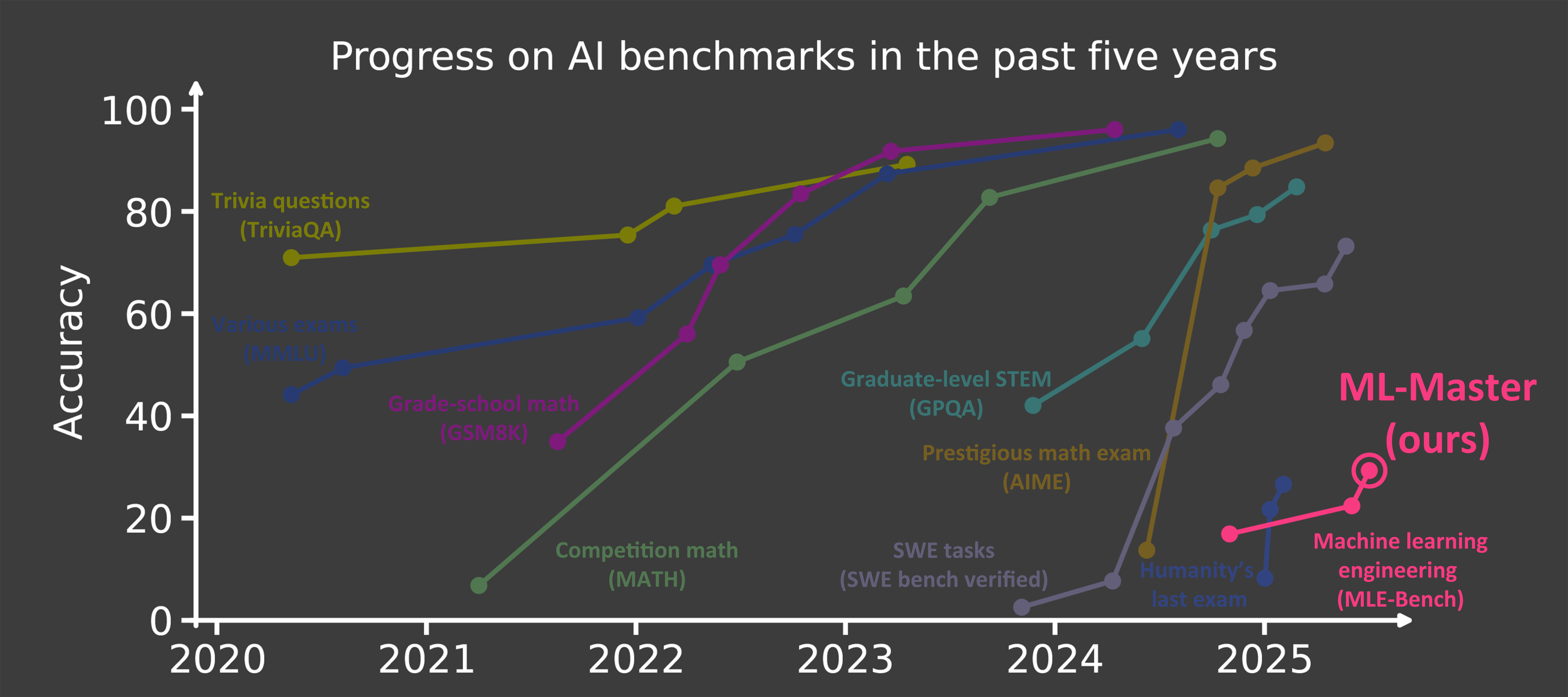}
    \caption{Positioned at the frontier of AI-for-AI progress, within a month, ML-Master boosted the performance of MLE-Bench, one of the most authoritative autonomous machine learning engineering benchmarks, by $30.8\%$ ($22.4\% \rightarrow 29.3\%$), highlighting the rapidly evolving path toward AI4AI.~\cite{secondhalf}.}
    \label{fig:second_half}
\end{figure}

We benched ML-Master on the widely recognized MLE-Bench~\cite{mlebench}, 
achieving state-of-the-art results that significantly surpass existing methods across multiple evaluation metrics. 
MLE-Bench, introduced by OpenAI, is a comprehensive benchmark designed to evaluate systems on  challenging real-world machine learning tasks derived from Kaggle competitions. 
As shown in Figure~\ref{fig:pre_abstract}, we measure performance using the average medal rate, defined as the percentage of tasks where the method achieves Bronze, Silver, or Gold-level performance. 
ML-Master achieved an average medal rate of 29.3\%, substantially outperforming the strongest baseline, 
R\&D-Agent~\cite{rdagent}, 
which achieved a medal rate of 22.4\%.  
Furthermore, ML-Master excels particularly in tasks categorized as medium difficulty, attaining an impressive medal rate of 20.2\%, more than doubling the previous best result of 9.0\%,  demonstrating its superior capability in handling complex and challenging AI development scenarios. 
Notably, ML-Master accomplished this superior performance within a strict time constraint of only 12 hours, merely half of the 24-hour limit previously employed by baselines. 
Results on MLE-Bench underscore ML-Master's consistent superiority across various evaluation dimensions. 
This substantial margin highlights ML-Master's ability to handle complex AI development tasks with remarkable efficiency and accuracy, further establishing its superiority in AI4AI.

In summary, our contributions are as follows:
\begin{itemize}[left=1em]
    \item We propose ML-Master, a novel AI4AI agent that employs an adaptive memory mechanism to seamlessly integrate comprehensive exploration and analytical reasoning into a unified framework, inspired by the cognitive strategies of expert AI developers. 
    
    \item We achieve state-of-the-art performance on the MLE-Bench, attaining an average medal rate of 29.3\% and excelling particularly in complex, medium-difficulty tasks, where we more than double the previous best result with a medal rate of 20.2\%.
    \item Remarkably, ML-Master delivers this exceptional performance with less computational cost than previous methods, requiring only 12 hours–half the time limit set by earlier approaches.
\end{itemize}

\section{Related Work}

\newcommand{\cmark}{\color[HTML]{32CB00}\CheckmarkBold}
\newcommand{\xmark}{\color[HTML]{FE0000}\ding{55}}

\begin{table}[!t]
    \centering
    \caption{Comparison of ML-Master with existing AI4AI methods. \textit{Chain} indicates sequential exploration following a linear path. 
    \textit{Multi-chain} refers to multiple sequential paths that explore different directions. 
    \textit{Tree} means structured tree-based exploration that systematically navigates solution spaces. 
    \textit{Uncontrollable} indicates that reasoning capabilities cannot be effectively guided by the system, limiting adaptability. }
    \scalebox{1.0}{
    \begin{tabular}{l|cccccc}
        \toprule
        \textbf{Method} &
        \makecell{\bf Exploration \\ \bf Strategy} &
        \makecell{\bf Reasoning \\ \bf Enhancement} &
        \makecell{\bf Adaptive \\ \bf Memory} &
        \makecell{\bf Parallel \\ \bf Execution} \\
        \midrule
        MLAB~\cite{mlagentbench} & Chain & \xmark & \xmark & \xmark  \\
        OpenHands~\cite{openhands} & Chain & \xmark  & \xmark & \xmark  \\
        SELA~\cite{sela} & Tree \& Predefined & \xmark &  \xmark & \xmark  \\
        AIDE~\cite{aide} & Tree \& Greedy  & Uncontrollable &  \xmark & \xmark  \\
        Agent Laboratory~\cite{agentlaboratory} & Tree \& Greedy & Uncontrollable  & \xmark & \cmark  \\
        R\&D-Agent~\cite{rdagent} & Multi-chain \& Fusion & Uncontrollable & \xmark & \cmark  \\
        \midrule
        \rowcolor[HTML]{DAE8FC}\textbf{ML-Master (ours)} & \textbf{Tree \& Balanced} & \textbf{Steerable} &  \cmark & \cmark  \\
        \bottomrule
    \end{tabular}
    }
    \label{tab:comparison}
\end{table}

\paragraph{Automated machine learning (AutoML).}
AutoML is the study of automating the tasks of machine learning engineering. By streamlining model development through both heuristic and learning-based approaches, AutoML serves as an initial step toward the broader AI4AI vision. Before the advent of LLMs, AutoML research have mainly focused on the repetitive and labor-intensive aspects of machine learning, such as data preprocessing, model selection and parameter tuning.~\cite{autogluon,tpot,autosklearn2, ml-plan, autogluon-tabular, liu2020admm} 
For example, AutoGluon-Tabular~\cite{autogluon-tabular} automates emsemble learning to fit end-to-end pipeline on tabular data with minimal user inputs. MOSAIC~\cite{rakotoarison2019automated} combines tree search with Bayesian optimization to effectively search for optimal model architecture and hyperparameter. 
However, these works have all relied upon heuristic methods, lacking adaptability and generalization abilities. 
Other works have used learning-based methods to optimize hyperparameters and select model architecture~\cite{zoph2017neuralarchitecturesearch,liu2018darts,real2020automl,zoph2018learning,wu2019fbnet};
for example, Zoph and Le~\cite{zoph2017neuralarchitecturesearch} trained an RNN to search for optimal neural networks on the CIFAR-10 dataset, rivaling the performance of human-designed models.
However, these works still require human to pre-define the pipeline before training, and are often restricted to specific tasks and datasets. Overall, classical AutoML works remain constrained by predefined search spaces and static configurations, lacking the adaptability and capabilities for continuous learning.

\paragraph{LLMs and multi-agent systems for AI4AI.} 
Recent advancements in LLMs have unlocked in a paradigm shift, transitioning from human-led to AI-led AI development. 
In contrast to earlier AutoML systems, LLMs are capable of complex reasoning, knowledge-based judgment and code generation.~\cite{deepseekr1,zheng2023judging,yao2023react,hui2024qwen2}
These advanced capabilities enable LLM-based systems to act freely and self-improve with minimal human intervention, thereby representing a significant step toward AI4AI.
For example, early works like AutoML-GPT~\cite{zhang2023automl} and MLCopilot~\cite{mlcopilot} exploit LLM's strong capabilities through prompt engineering to automate the entire machine learning pipeline. 
Recent works have focused on LLM-based multi-agent systems (MAS): 
AutoKaggle~\cite{autokaggle} and Agent K~\cite{grosnit2024agentk} designs a fully-automated multi-agent framework modeled after human engineering process to allow LLM agents to compete in Kaggle competitions; 
MLAgentBench~\cite{mlagentbench} and MLZero~\cite{fang2025mlzero} introduces external tools and memory to enhance LLM agents in AI research and development. 
Dolphin~\cite{dophin} introduces retrieval-augmented generation and an iterative refinement strategy to enhance the idea proposal process. Some recent works such as
Agent Laboratory~\cite{agentlaboratory}, NovelSeek~\cite{novelseek}, the AI Scientist~\cite{ai_scientist} and AI-Researcher~\cite{airesearcher}
have gone further to automate the entire AI research process from idea proposal to code implementation, validation and paper generation.
While powerful, due to limitations in pipeline design, LLM-based multi-agent systems often struggle with insufficient exploration, a significant drawback considering the immense scale and complexity of machine learning problems.

\paragraph{Self-evolving AI.}
Self-evolving AI enables itself to autonomously acquire, refine, and learn from self-generated or personalized experiences, continuously improving their reasoning and adaptability without heavy reliance on human intervention. 
Current research in self-evolving AI spans multiple domains and approaches. 
For instance, DeepSeek-R1-Zero~\cite{deepseekr1} enhances reasoning via iterative self-verification and extended chain-of-thought generation, STaR~\cite{star} bootstraps reasoning by leveraging its own outputs for self-training, and RStar-Math~\cite{rstarmath} enables small LLMs to master math reasoning through self-evolved deep-thinking strategies.
In addition, works in autonomous agents showcase self-evolving by integrating experience accumulation, for example, 
Reflexion~\cite{reflexion} employs verbal reinforcement learning to enable self-reflective improvement, 
while EvoMac~\cite{evomac} demonstrates self-evolving multi-agent collaboration for software development. 
Meanwhile, some studies on tool-using systems, such as Alita~\cite{alita}, WebRL~\cite{webrl}, AgentGym~\cite{agentgym}, also exemplify self-evolution by autonomously creating, refining, and adapting external resources to enhance their capabilities.
These endeavors demonstrate that AI can progressively achieve higher levels of autonomy and generalization through self-evolving mechanisms without reliance on manual engineering or external supervision.
In the endgame of AI4AI domain, AI systems can autonomously design and optimize other AI systems while simultaneously refining their own strategies, achieving continuous capability enhancement and creating increasingly advanced AI through self-evolution.

\section{Methodology}

\begin{figure}[!t]
    \centering
    \includegraphics[width=1\linewidth]{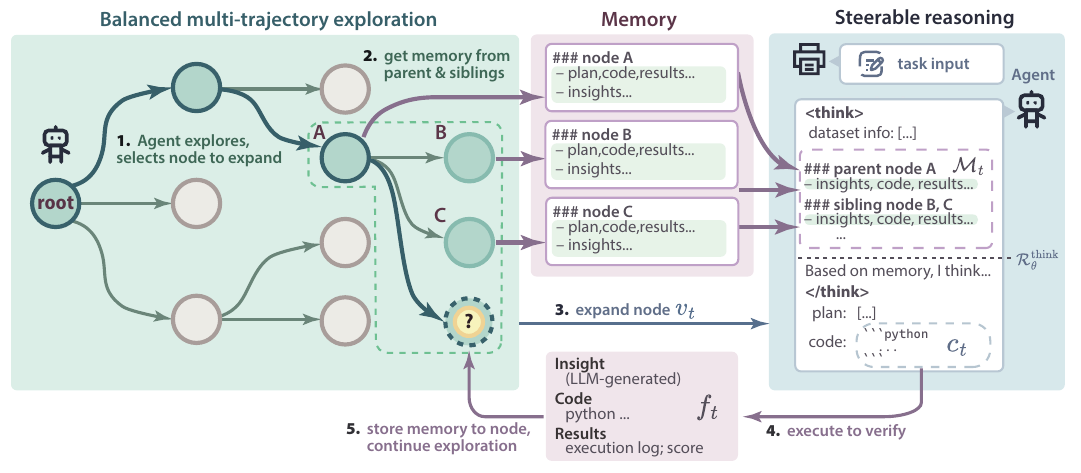}
    \caption{An overview of ML-Master's two modules: \textbf{balanced multi-trajectory exploration} and \textbf{steerable reasoning}. Several draft solutions (nodes) are initialized at the start of each run. In each exploration step, an LLM agent is prompted to either improve or debug a previous solution, expanding from different draft nodes following an MCTS-inspired approach, enhanced with parallelism to ensure efficiency. Memories and insights from all explored branches are then fed into the LLM agent's reasoning process, leading to more steerable reasoning and higher performance.}
    \label{fig:pipeline}
\end{figure}

The development of robust AI systems within the AI4AI paradigm necessitates a unified framework integrating exploration and reasoning. 
Exploration enables the agent to traverse diverse solution paths and adapt to complex uncertainties, while reasoning empowers the agent to interpret, evaluate, and synthesize information. 
However, in isolation, each is limited: 
exploration alone may devolve into inefficient trial-and-error, 
while reasoning alone risks analytical stagnation when confined to prior knowledge. 
Therefore, the integration of exploration and reasoning is essential for autonomous agents to achieve both depth and breadth in AI development.

To realize this integration, we propose ML-Master, a tightly coupled iterative agent combining steerable reasoning module and balanced multi-trajectory exploration module through a adaptive memory mechanism.  
As shown in Figure~\ref{fig:pipeline}, 
balanced multi-trajectory exploration (\S~\ref{sec: multi-trace exploration}) enables ML-Master to generate and evaluates multiple solution trajectories in parallel, enriching the reasoning process with diverse empirical insights. 
This parallel exploration actively produces diverse empirical insights and execution feedback, which are adaptively captured and structured into a concise memory. 
Rather than overwhelming the reasoning process with extensive and redundant historical data, this memory strategically retains only the most relevant and actionable insights derived from exploration. 
Concurrently, steerable reasoning (\S~\ref{sec: steerable reasoning}) embeds adaptive memory into the reasoning process,  ensuring precise, reliable, and controlled analytical capabilities while significantly reducing hallucinations and erroneous interpretations. 
The reasoning module thus effectively interprets exploration outcomes, 
with insights derived from this reasoning process directly informing and guiding subsequent exploration. 
Together, these two modules form a closed-loop system, where exploration continuously enriches reasoning with empirical insights, and reasoning systematically directs exploration toward promising trajectories. 
This iterative interplay, guided by the adaptive memory mechanism, enables ML-Master to progressively refine its solutions, robustly navigate complex problem spaces, and achieve superior performance in AI4AI.

\subsection{Balanced Multi-Trajectory Exploration}
\label{sec: multi-trace exploration}

\begin{figure}[!t]
    \centering
    \includegraphics[width=0.99\linewidth]{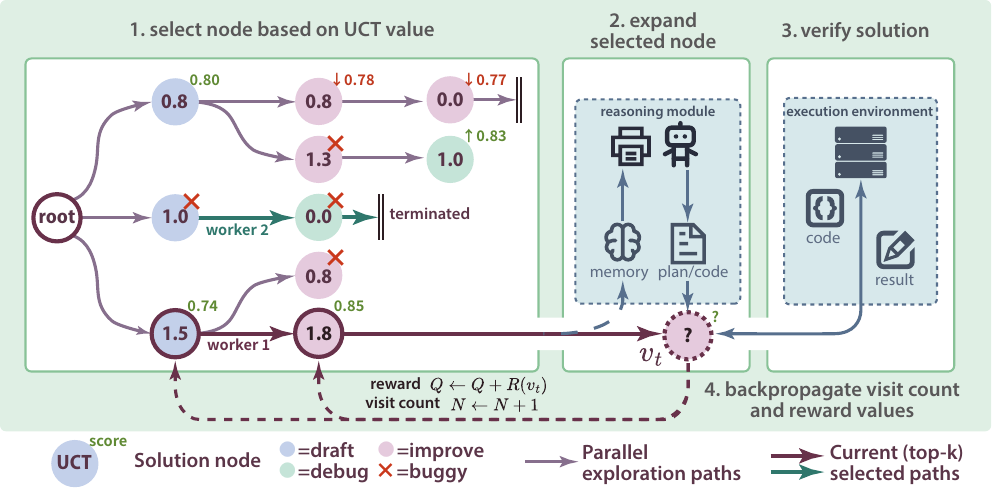}
    \caption{The balanced multi-trajectory exploration pipeline operates through a tree-guided exploration with parallel search. It combines MCTS-inspired tree search with parallel exploration: (1) selection traverses from root to leaf using UCT criterion; (2) expansion generates child nodes through the steerable reasoning module; (3) verification evaluates solutions in the execution environment; (4) backpropagation propagates rewards and visit counts upward. Multiple workers explore different branches asynchronously, with top-k nodes serving as entry points for deeper parallel search.}
    \label{fig:multi-trace}
\end{figure}

In this section, we present the balanced multi-trajectory exploration module in ML-Master, which is designed to efficiently explore multiple solution trajectories in parallel. 
Given the complexity and scale of the search space in AI4AI tasks, it is essential to balance exploration breadth and depth effectively. 
As illustrated in Figure~\ref{fig:multi-trace}, our balanced multi-trajectory exploration module adopts a structured, tree-based approach to guide exploration strategically, inspired by Monte Carlo Tree Search (MCTS). 
Specifically, balanced multi-trajectory exploration consists of two complementary components:
(1) Tree-guided exploration, 
which reformulates the AI development process as an iterative exploration of potential solutions, using MCTS to efficiently navigate the solution space by constructing and expanding a search tree. This allows ML-Master to prioritize under-explored solution paths and dynamically adjust exploration efforts. 
(2) Parallel search, 
which allows for concurrent exploration of multiple branches within the search tree, significantly improving both the efficiency and scalability of the search process. 
In the following subsections, we detail these two components and explain how their integration enables comprehensive and efficient exploration, ultimately enriching the reasoning process with diverse empirical insights.

\subsubsection{Tree-guided exploration}

To comprehensively explore the vast and complex solution space inherent in AI development tasks, 
we propose a novel formulation that explicitly models the AI development process as Monte Carlo Tree Search (MCTS). 
Specifically, ML-Master 
constructs and expands a structured search tree, where each node represents a distinct solution state, and edges correspond to specific refinement actions. 
By leveraging a tree-based structure, ML-Master can efficiently manage and prioritize exploration efforts, ensuring comprehensive coverage of diverse solution trajectories while maintaining computational efficiency.

Specifically, the exploration process involves four key phases: 
(a) selection, where ML-Master employs a novel context-aware criterion to efficiently prioritize promising yet under-explored solution states; 
(b) expansion, where ML-Master takes specialized refinement actions uniquely tailored to AI development tasks; 
(c) verification,  where ML-Master adopts a reward function to accurately and efficiently assess the quality of candidate solutions, and 
(d) backpropagation, where ML-Master propagates structured evaluation feedback through the search tree, dynamically guiding subsequent exploration decisions. 
They are executed iteratively and in parallel across multiple solution paths.

\paragraph{Selection.} 
The selection process begins at the root node and recursively selects the child node 
with the highest Upper Confidence Bound for Trees (UCT) value until it reaches either a leaf node or a node that is not fully expanded.
Formally, the UCT value for node $v$ is defined as:
\begin{equation}
\text{UCT}(v) = \frac{Q_v}{N_v} + C \cdot \sqrt{\frac{\ln N_\text{parent}}{N_v}},
\end{equation}
where $Q_v$ is the total reward of node $v$, $N_v$ is the visit count of node $v$, $N_\text{parent}$ is the total number of visits to the parent node, and $C$ is a constant controlling the exploration-exploitation trade-off.
Nodes with higher UCT values represent promising yet under-explored solution paths, thus guiding the exploration towards potentially valuable regions of the solution space.

During the selection phase of MCTS, a node is treated as terminal and excluded from further expansion if it satisfies any of the following stopping conditions:

First, we define an \textbf{improvement-based termination} criterion to eliminate nodes that show persistent stagnation. Let $\Delta_i$ denote the relative improvement over the best ancestor node on the current path after the $i$-th improve operation. If the number of failed improvements—i.e., those not exceeding a predefined threshold $t$—exceeds a tolerance level $\tau_{\text{improve}}$, the node is considered terminal:
\begin{equation}
\label{eq:improve-terminate}
    \sum_{i=1}^{\mathrm{K}} \mathbb{I}[\Delta_i < t] > \tau_{\text{improve}},
\end{equation}
where $\mathbb{I}[\cdot]$ is the indicator function and $\mathrm{K}$ is the number of Improve attempts made at the node.

Second, we enforce a \textbf{debugging depth constraint} to prevent the search from persistently attempting to fix nodes.If the number of consecutive debug operations up to the current node exceeds $\tau_{\text{debug}}$, the node is marked as terminal.

These two constraints jointly act as a pruning mechanism to suppress unproductive search trajectories and allocate computational resources toward more promising regions of the solution space.

\paragraph{Expansion.} The expansion process starts from the selected node and applies three types of actions to generate new child nodes. These actions—Draft, Debug, and Improve—are designed to guide the search toward higher-quality solutions by addressing different aspects of code refinement and generation:
\begin{itemize}[left=1em]
    \item \textbf{Draft}: The Draft action generates an initial, runnable code solution for node. Its primary function is to produce a basic implementation that satisfies the task requirements, serving as a starting point for further refinement.
    \item \textbf{Debug}: The Debug action focuses on identifying and correcting the errors in the current code. Its main role is to ensure that the node’s code compiles and executes correctly before any further enhancements.
    \item \textbf{Improve}: The Improve action enhances the quality of functionally correct code by tune data preprocessing, model architecture, or optimization approaches. Its purpose is to generate a node that incorporates these refinements and achieves quantifiable performance gains.
\end{itemize}

For each node $v$, we provide a formal specification of the decision rule that determines which action to apply at each step. 
The selection is governed by the following condition:
\begin{itemize}[left=1em]
\item If there is no existing solution for the task in node $v$, the next step is taking the action \textbf{Draft} to draft a new one.
\item If node $v$ contains a solution that still has bugs, and debugging is not yet complete, the next step is taking the action \textbf{Debug} to identify and fix the remaining issues.
\item If the solution in node $v$ is currently bug-free, but further improvement is still needed, the next step is taking the action \textbf{Improve} to enhance its performance.
\end{itemize}

To allow iterative progress, we define stopping conditions for both the debugging and improving phases.
The debugging process ends when either a correct solution has been produced or the number of debugging attempts exceeds a predefined limit.
The improving process terminates once no further progress can be made. 
The design supports iterative refinement while bounding the number of attempts, striking a balance between thoroughness and efficiency.

\paragraph{Verification.} The verification process evaluates the quality of the newly expanded node $v$ by computing a reward signal that reflects its effectiveness in addressing the task objectives. The reward function $R(v)$ is defined as:
\begin{equation}
R(v) = 
\begin{cases}
-1, & \text{if } \mathcal{D}(v) \\
r_q(v) + r_d(v) + r_s(v), & \text{otherwise,}
\end{cases}
\end{equation}
where $\mathcal{D}(v)$ determines whether node $v$ contains defects and the reward components are defined as follows:
\begin{itemize}[left=1em]
    \item Quality reward $r_q(v)$: Indicates whether the solution represented by $v$ improves upon the best evaluation metric observed so far. Formally:
    \begin{equation}
        r_q(v) = 
            \begin{cases}
            1, & \text{if } M(v) > M^* \\
            0, & \text{otherwise.}
            \end{cases}
    \end{equation}
    Here, $M(v)$ denotes the evaluation metric used to assess the solution quality at node $v$, and $M^*$ is the best score observed so far during the search.
    \item Debugging reward $r_d(v)$: Reflects whether the transition from the parent node to $v$ successfully eliminates a previously identified fault. That is, $r_d(v) = 1$ if the fault present in the parent node is resolved in $v$; otherwise, $r_d(v) = 0$.
    \item Structural improvement reward $r_s(v)$:This term reflects whether the node $v$ represents the successful completion of an improvement process. Specifically, if $v$ satisfies a predefined stopping criterion $\mathcal{T}_i(v)$, which signals that the stopping condition for the improvement process has been met, then $r_s(v) = 1$; otherwise, $r_s(v) = 0$.
\end{itemize}
\paragraph{Backpropagation.} After the verification phase, the obtained reward is propagated back along the path from the expanded node to the root. During this backpropagation process, each node along the path updates its visit count $N$ and accumulated reward $Q$ accordingly. 

\subsubsection{Parallel search}
To effectively scale Monte Carlo Tree Search (MCTS) to parallel environments and large search spaces, we aim to design a search framework that enables asynchronous exploration across promising subregions of the tree while preserving the core principles of MCTS.

To this end, we introduce an asynchronous branch-parallel MCTS strategy. The search begins with all workers jointly expanding the root node in parallel. Once the root's children are fully expanded, the top-k nodes with the highest UCT values are selected as new entry points for deeper search. Each selected node then initiates an independent search thread, allowing selection, expansion, verification, and backpropagation to proceed asynchronously and without cross-thread interference.

When a thread completes its search within a branch, it returns to the root and selects the best available child node—based on UCT score—from among those not currently being explored by other threads. This mechanism ensures that parallel exploration proceeds efficiently without duplication, and that computational resources are dynamically reallocated to promising yet unoccupied regions of the search tree.

The proposed scheme enables broad and adaptive exploration of the search space while maintaining consistency with the MCTS framework. It is particularly well-suited to tasks with large branching factors or non-uniform value distributions among subtrees.

\subsection{Steerable Reasoning}
\label{sec: steerable reasoning}

\begin{figure}
    \centering
    \includegraphics[width=0.90\linewidth]{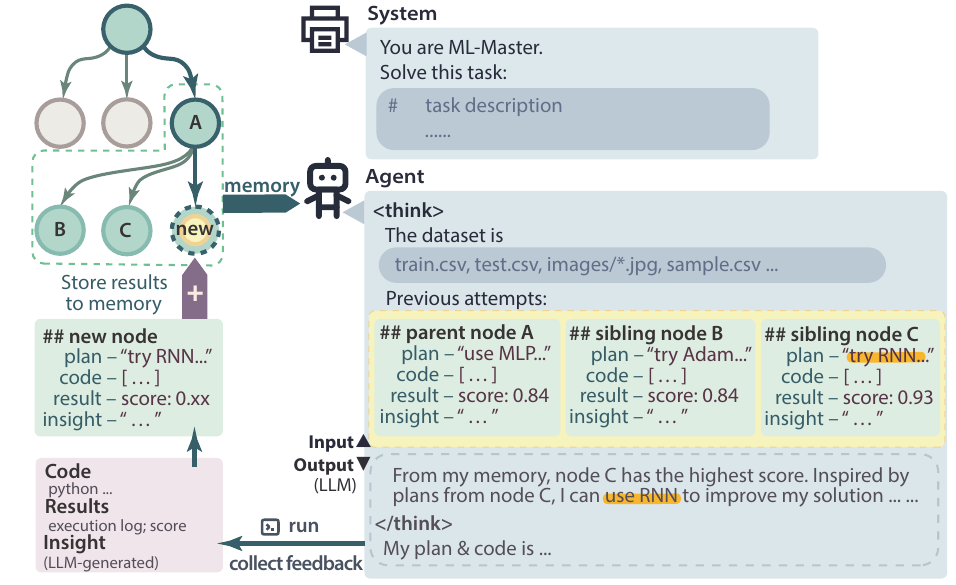} 
    \caption{Steerable reasoning with adaptive memory. The memory aggregates insights from the immediate parent node and parallel sibling nodes at the same exploration depth, containing execution results, code snippets, and performance scores. During reasoning, this curated memory is explicitly embedded into the LLM's "\texttt{think}" component, enabling contextually grounded decision-making while avoiding information overload. The generated plan and code are executed through an interpreter, producing execution logs, submission files, and performance scores that are saved back to the search tree, forming a closed-loop learning system.}
    \label{fig:reasoning}
\end{figure}

Steerable Reasoning is essential to any AI4AI that seeks to improve through self-guided iteration. 
In ML-Master, we significantly enhance the reasoning capabilities of the advanced reasoning model (Deepseek-R1~\cite{deepseekr1}) by explicitly embedding contextual memory directly into the reasoning ("{\texttt{think}}") component, rather than into the instruction component. 
Specifically, it adaptively leverages only the immediate previous reasoning node and parallel sibling nodes within the same exploration depth. 
This adaptive memory mechanism ensures precise, coherent, and contextually grounded reasoning, effectively reducing hallucinations and redundancy. 
By dynamically extracting the most valuable analytical insights and integrating concise execution feedback, ML-Master continuously learns from past experiences, enabling robust analysis, effective debugging, and informed decision-making, thereby substantially improving the reliability and performance of AI4AI.

\textbf{Reasoning process and insight extraction.} 
Formally, at each exploration node $t$, ML-Master performs reasoning based on its current task input $x$ and contextual memory from previous explorations. The complete reasoning process $c_t$ is generated as:
\begin{equation}
c_t = \mathcal{R}_\theta^{think}(x , \mathcal{M}_t),
\end{equation}
where $\mathcal{R}_\theta^{think}$ explicitly indicates the reasoning ("\texttt{think}") component of the reasoning LLM, conditioned on the input $x$ and the memory $\mathcal{M}_t$ (defined in Eq.~\ref{eq: memory}). 
The reasoning process $c_t$ encompasses the complete step-by-step analytical thinking, including problem analysis, strategy formulation, decision rationale, and final solution.

From this comprehensive reasoning process, we systematically extract the most salient insights:
\begin{equation}
    r_t = \varepsilon(c_t),
\end{equation}
where $\varepsilon(\cdot)$ extracts key analytical insights, identified patterns, debugging strategies, and improvement directions from the full reasoning trace $c_t$. 
This extraction process is crucial as it transforms verbose reasoning into concise, actionable knowledge that can effectively guide future exploration without overwhelming the reasoning process.

\textbf{Execution feedback collection.} 
In parallel, execution feedback $f_t$ is collected through the interpreter during each expansion step, capturing: 
(1) performance metrics from model evaluation on validation data, 
(2) execution logs from code compilation and runtime, and 
(3) error diagnostics when solutions fail to execute properly. This empirical feedback provides concrete evidence of solution quality and identifies specific areas requiring improvement.

\textbf{Adaptive memory construction.} 
To facilitate controlled reasoning, we construct an adaptive memory $\mathcal{M}_t$ that strategically combines distilled insights and execution feedback while avoiding redundant trajectories:
\begin{equation}
\mathcal{M}_t = \left\{(r_{t-1}, f_{t-1})\right\} \cup \left\{(r_{t}^{(s)}, f_{t}^{(s)}) \;\middle|\; s \in \mathcal{S}_{t}\right\},
\label{eq: memory}
\end{equation}
where $(r_{t-1}, f_{t-1})$ denotes reasoning insights and execution feedback from the immediately preceding node $t-1$ within the current exploration branch, ensuring logical continuity and progressive refinement. 
$\mathcal{S}_{t}$ denotes the set of sibling nodes at the same exploration depth within parallel branches, with each sibling node $s$ providing alternative reasoning contexts $(r_{t}^{(s)}, f_{t}^{(s)})$ that offer contrastive perspectives on the same task. 
The inclusion of sibling-node information introduces contrastive signals, allowing the model to recognize and avoid producing reasoning paths that mirror those already explored in parallel, thereby promoting diversity and preventing redundant exploration. Meanwhile, the direct lineage from the previous node ensures logical continuity within each exploration branch. 

By explicitly embedding this carefully curated contextual memory into the LLM's reasoning component, ML-Master achieves steerable reasoning that is both contextually coherent and diversified across parallel exploration paths. 
This controlled integration significantly reduces common pitfalls such as hallucinations, redundant reasoning, and convergence to suboptimal solutions, thereby substantially improving the reliability and effectiveness of AI4AI.

\subsection{Discussions}

\textbf{Rationality and Advantages of Integrating Exploration and Reasoning.} 
The design of ML-Master is grounded in cognitive principles that characterize AI expert problem-solving behavior. 
AI expertise demonstrates that effective problem-solving emerges from the synergistic interaction between exploration and reasoning. 
ML-Master realize this principle by creating a bidirectional information flow: exploration generates empirical evidence that enriches reasoning context, while reasoning provides strategic guidance that focuses exploration efforts. 
This integration addresses a fundamental limitation in existing AI4AI methods, where exploration and reasoning operate in isolation, leading to either inefficient trial-and-error exploration or analytically sound but empirically ungrounded solutions. 
The adaptive memory mechanism serves as the critical bridge between these processes, selectively retaining insights from exploration trajectories while avoiding information overload that can degrade reasoning performance. 
By curating contextual information from both direct lineage (parent nodes) and sibling nodes, ML-Master maintains both coherence and diversity in its reasoning process. 
This unified approach enables ML-Master to navigate complex AI tasks effectively, scaling human-like expertise to accelerate AI4AI.

\textbf{Comparison with Existing AI4AI Methods.} 
Table~\ref{tab:comparison} systematically compares ML-Master against existing AI4AI methods across four critical architectural dimensions, revealing fundamental limitations in current approaches and highlighting ML-Master's architectural innovations.
Early systems such as MLAB~\cite{mlagentbench} and OpenHands~\cite{openhands} employ straightforward chain-based exploration following linear solution paths, but lack reasoning enhancement and adaptive memory capabilities, making them unable to learn from accumulated experience or strategically guide exploration. 
More sophisticated approaches like SELA~\cite{sela} introduce tree-based search mechanisms but fail to leverage advanced reasoning models, treating exploration and reasoning as independent processes. 
AIDE~\cite{aide}, Agent Laboratory~\cite{agentlaboratory}, and R\&D-Agent~\cite{rdagent} represent progress by incorporating reasoning capabilities, yet suffer from critical flaws: their reasoning processes remain uncontrollable and cannot be effectively steered by exploration outcomes, while their memory mechanisms are fixed, preventing adaptive learning from exploration history. 
This disconnect leads to unreliable outputs, hallucinations, and suboptimal performance, particularly in complex tasks where strategic coordination becomes essential. 
In contrast,  ML-Master integrates exploration and reasoning into a unified iterative framework via an adaptive memory mechanism, enabling the reasoning process to be directly informed by exploration outcomes while providing strategic guidance for subsequent exploration steps. 
This architectural advancement, combined with parallel execution capabilities, enables ML-Master to achieve both computational efficiency and exploration quality—a balance that existing methods struggle to maintain due to their fragmented designs.

\section{Experiment}

\subsection{Experiment Setup}
\textbf{Settings. }Our ML-Master agent is tested on MLE-Bench~\cite{mlebench}, a diverse and realistic benchmark introduced by OpenAI to assess AI agents on end-to-end machine learning engineering tasks.
The following configuration is informed by the algorithmic components introduced earlier.
The predefined threshold $t$ is used to determine whether an improve action is successful. To tolerate occasional stagnation, the process allows up to $\tau_{\text{improve}}$ consecutive failed improve attempts. 
In our experiments, we use $t=0.001$ and $\tau_{\text{improve}}=3$.
To prevent persistent attempts at fixing nodes, a debug depth constraint $\tau_{\text{debug}}=20$ is enforced: nodes exceeding this limit are marked terminal. Additionally, each debug action sequence ends once it reaches a depth of 3, thereby enabling iterative progress.
DeepSeek-R1-0120~\cite{deepseek-r1} is employed to generate plans and code for the corresponding actions.
The overall search procedure is executed in parallel at the MCTS level, with a parallelism degree of 3.

\textbf{Environment. } We try our best to ensure the same testing environment with AIDE. However, due to the limitation of hardware, there are still some differences. In our experiments, each agent is equipped with 36 AMD EPYC vCPUs and one NVIDIA A100 Tensor Core GPU. Every three agent share 512GB memory and 1TB SSD to produce submissions and any intermediate files. The total time of a task is set to 12 hours. Overall, our testing environment is slightly inferior to the one reported by MLE-Bench. 

\textbf{Baselines.} To provide a comprehensive comparison, we compare ML-Master with OpenHands~\cite{openhands}, MLAB~\cite{mlagentbench}, AIDE~\cite{aide} and R\&D-Agent~\cite{rdagent}. Due to the expensive cost of running complete MLE-Bench, we use some results reported by MLE-Bench itself.
Additionally, we run AIDE, which achieved the best performance on MLE-Bench before our method, using Deepseek-R1-0120~\cite{deepseek-r1} to provide a fair comparison between ML-Master and AIDE.

\subsection{Main Results}
\begin{table}[!t]
\centering
\caption{ML-Master outperforms all baselines on all evaluation dimensions defined by MLE-Bench. The results of MLAB, OpenHands, AIDE (gpt-4o-2024-08-06 and o1-preview) and R\&D-Agent are report by official MLE-Bench. Results for ML-Master are averaged over 3 runs with different random seeds and reported as mean ± one standard error of the mean (SEM). The top-performing model is shown in \textbf{bold}. * indicates that only a single run is conducted due to time and resource constraints.}
\label{tab:mcts-result}
\small
\begin{tabularx}{\textwidth}{lYYYYYY}
\toprule
\textbf{Agent}  & \textbf{Valid} & \textbf{Above} & \textbf{Bronze} & \textbf{Silver} & \textbf{Gold} & \textbf{Any} \\
 & \textbf{Submission} & \textbf{Median} & & & & \textbf{Medal} \\
 &($\%$)  & ($\%$) & ($\%$) & ($\%$) & ($\%$) & ($\%$) \\
\midrule
\multicolumn{7}{l}{\textbf{MLAB}~\cite{mlagentbench}} \\
\midrule
gpt-4o-2024-08-06 & 44.3 ± 2.6 & 1.9 ± 0.7 & 0.0 ± 0.0 & 0.0 ± 0.0 & 0.8 ± 0.5 &  0.8 ± 0.5\\
\midrule
\multicolumn{7}{l}{\textbf{OpenHands}~\cite{openhands}} \\
\midrule
gpt-4o-2024-08-06 & 52.0 ± 3.3 & 7.1 ± 1.7 & 0.4 ± 0.4 & 1.3 ± 0.8 & 2.7 ± 1.1 &  4.4 ± 1.4\\
\midrule
\multicolumn{7}{l}{\textbf{AIDE}~\cite{aide}} \\
\midrule
gpt-4o-2024-08-06 & 54.9 ± 1.0 & 14.4 ± 0.7 & 1.6 ± 0.2 & 2.2 ± 0.3 & 5.0 ± 0.4 &  8.7 ± 0.5\\
o1-preview & 82.8 ± 1.1 & 29.4 ± 1.3 & 3.4 ± 0.5 & 4.1 ± 0.6 & 9.4 ± 0.8 &  16.9 ± 1.1\\
Deepseek-R1* & 78.6 ± 0.0 & 34.6 ± 0.0 & 2.7 ± 0.0 & 4.0 ± 0.0 & 8.0 ± 0.0 &  14.7 ± 0.0\\
\midrule
\multicolumn{7}{l}{\textbf{R\&D-Agent}~\cite{rdagent}} \\
\midrule
o1-preview & 86.1 ± 1.1 & 32.8 ± 1.2 & 3.5 ± 0.5 & 4.5 ±  0.5& 14.4 ± 0.5&  22.4 ± 0.5 \\
\midrule
\multicolumn{7}{l}{\textbf{ML-Master}} \\
\midrule
\rowcolor[HTML]{DAE8FC}Deepseek-R1 & \textbf{93.3 ± 1.3} & \textbf{44.9 ± 1.2}&\textbf{4.4 ± 0.9} & \textbf{7.6 ± 0.4} & \textbf{17.3 ± 0.8} & \textbf{29.3 ±  0.8} \\
\bottomrule
\end{tabularx}
\end{table}

We evaluate our ML-Master on complete MLE-Bench among 75 machine learning tasks. ML-Master is compared against 4 methods driven by 3 models. We use the same evaluation metric as MLE-Bench. Results are shown in Table~\ref{tab:mcts-result} and demonstrate that: 
\begin{itemize}[left=1em]
    \item \textbf{ML-Master receives a medal in $\mathbf{29.3\%}$ of the machine learning tasks, with $\mathbf{17.3\%}$ achieving gold medals.} This indicates that ML-Master exhibits a remarkably high upper bound in performance and exceeds most human machine learning researchers when addressing specific machine learning tasks.
    \item \textbf{ML-Master makes a valid submission on $\mathbf{93.3\%}$ tasks and achieves performance superior to more than half of human submissions in $\mathbf{44.9\%}$ of the tasks. } This indicates that ML-Master has a very high lower bound when handling a wide variety of machine learning tasks of various difficulty, demonstrating its ability to tackle diverse machine learning challenges.
\end{itemize}

\begin{table}[!t]
\centering
\caption{Percentage of achieving any medals across different machine learning task complexity levels. ML-Master outperforms all baselines at each complexity level. Results for ML-Master are averaged over 3 runs with different random seeds and reported as mean ± one standard error of the mean (SEM). The top-performing model is shown in \textbf{bold}. * indicates that only a single run is conducted due to time and resource constraints.}
\label{tab:complexity-result}
\small
\begin{tabularx}{\textwidth}{lYYYY}
\toprule
\textbf{Agent}  & \textbf{Low(\%)} & \textbf{Medium(\%)} & \textbf{High(\%)} & \textbf{Average(\%)} \\
\midrule
\multicolumn{5}{l}{\textbf{MLAB~\cite{mlagentbench}}} \\
\midrule
gpt-4o-2024-08-06 & 4.2   ±  1.5 & 0.0   ±  0.0 & 0.0 ± 0.0 & 1.3   ±  0.5\\
\midrule
\multicolumn{5}{l}{\textbf{OpenHands~\cite{openhands}}} \\
\midrule
gpt-4o-2024-08-06 & 11.5  ±  3.4 & 2.2   ±  1.3 & 1.9   ±  1.9 & 5.1   ±  1.3\\
\midrule
\multicolumn{5}{l}{\textbf{AIDE~\cite{aide}}} \\
\midrule
gpt-4o-2024-08-06 & 19.0    ±  1.3 & 3.2   ±  0.5  & 5.6   ±  1.0  & 8.6   ±  0.5\\
o1-preview & 34.3   ±  2.4 & 8.8    ±  1.1  & 10.0   ±  1.9  & 16.9   ±  1.1\\
Deepseek-R1* & 27.3 ± 0.0 & 7.9 ± 0.0 & 13.3 ± 0.0 & 14.7 ± 0.0 \\
\midrule
\multicolumn{5}{l}{\textbf{R\&D-Agent~\cite{rdagent}}} \\
\midrule
o1-preview & 48.2 ±  1.1 & 8.9  ±  1.0 & 18.7  ±  1.3 & 22.4 ±  0.5\\
\midrule
\multicolumn{5}{l}{\textbf{ML-Master}} \\
\midrule
\rowcolor[HTML]{DAE8FC}Deepseek-R1  & \textbf{48.5  ±  1.5 } & \textbf{20.2  ±  2.3 }& \textbf{24.4  ±  2.2} & \textbf{29.3  ±   0.8}  \\
\bottomrule
\end{tabularx}
\end{table}

\begin{figure}[!t]
    \centering
    \begin{minipage}[c]{0.48\textwidth}
    \centering
    \includegraphics[width=\linewidth]{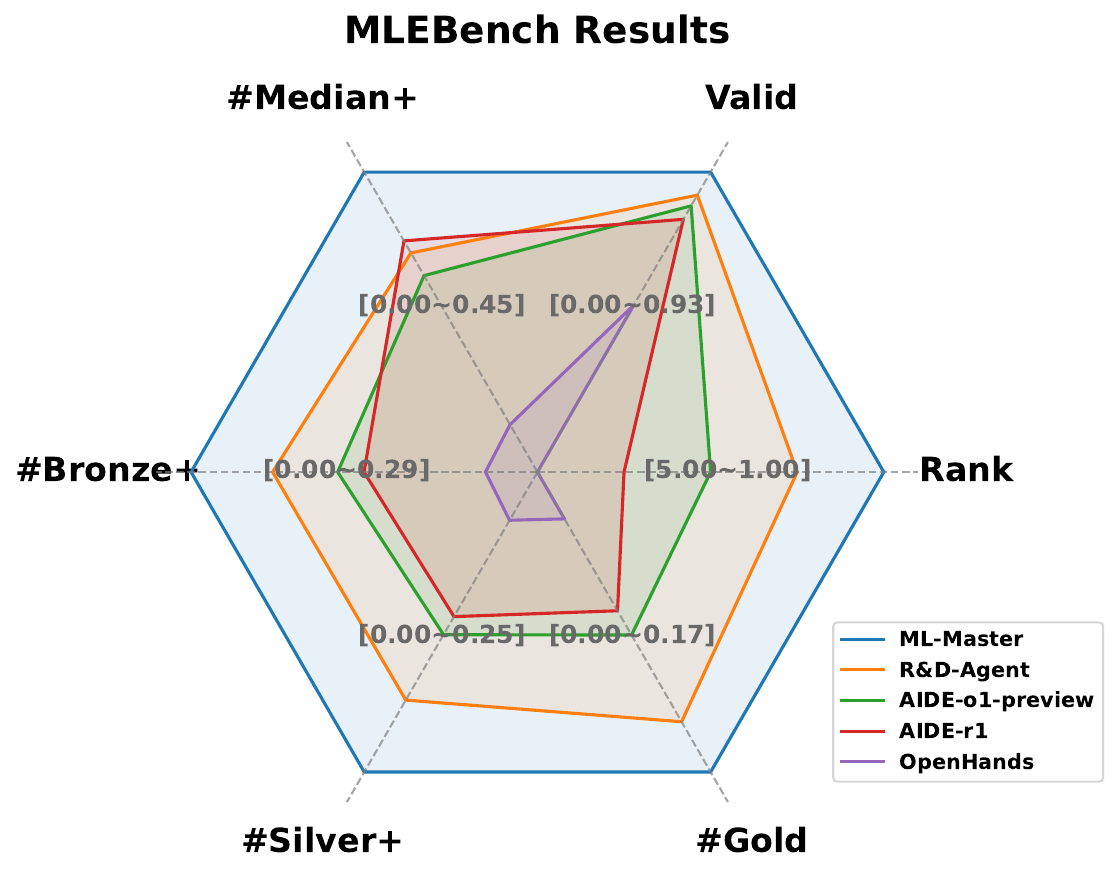}
    \caption{Performance of OpenHands, AIDE-r1, AIDE-o1-preview, R\&D-Agent and ML-Master. The plus notation indicates an equal or better result to the threshold. ML-Master performs better on all dimensions.}
    \label{fig:radar}
    \end{minipage}
    \hfill
    \begin{minipage}[c]{0.48\textwidth}
    \centering
    \vspace{10mm}
    \includegraphics[width=\linewidth]{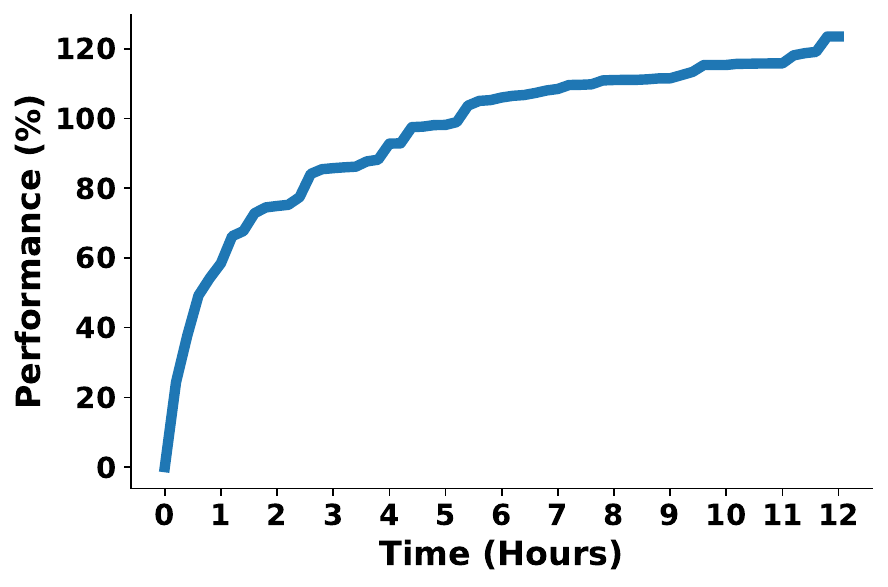}
    \vspace{-5mm}
    \caption{ML-Master's performance improves over time. Here, performance refers to the percentage improvement of the best version of the solution up to a certain point in time, compared to its initial version.}
    \label{fig:scaling}
    \end{minipage}
\end{figure}

\textbf{ML-Master outperforms all baselines on every evaluation dimension defined by MLE-Bench.} To better show the ability of ML-Master, we replace Bronze and Silver using Bronze+, Silver+ respectively. The plus notation indicates a result equal to or better than the threshold. For example, Silver+ represents the rate of reaching either a silver or gold medal. The results are shown in Figure~\ref{fig:radar}. We see that ML-Master outperforms other methods across all evaluation metrics.

\textbf{ML-Master is good at handling more complex machine learning tasks. }We further regroup the tasks by their complexity level as given in MLE-Bench, and calculate the medal rate of tasks in each complexity level. The results are shown in Table~\ref{tab:complexity-result}. We see that ML-Master achieves a $20.2\%$ medal rate on medium complexity tasks and $24.4\%$ medal rate on high complexity tasks, which benefits from the continuous reasoning and exploring process of ML-Master.

\subsection{Analysis}
\textbf{ML-Master continues improving its solution over time.} In Figure~\ref{fig:scaling},  we show how the solution given by ML-Master evolves over time. The vertical axis represents the improvement (in percentage) of the best version provided by ML-Master up to a certain point in time, compared with its initial version. The horizontal axis represents the iteration time. As the iteration time increases, ML-Master produces increasingly better solutions, demonstrating the effectiveness of its self-exploration and iterative reasoning process.

\textbf{Ablations.} We are actively conducting further ablation experiments on ML-Master and will report them in updated versions of this report.


\section{Conclusions}
In this paper, we introduced ML-Master, a novel AI4AI agent designed to seamlessly integrate exploration and reasoning into a unified framework. By employing an adaptive memory mechanism, ML-Master effectively combines parallel solution exploration with reasoning, contributing to significant advancements in AI4AI.

Our extensive evaluation on the MLE-Bench demonstrates the effectiveness of ML-Master, surpassing existing AI4AI methods. Notably, ML-Master achieved an average medal rate of 29.3\%, outperforming previous state-of-the-art methods. 
In articular, in medium-difficulty tasks, it attains an impressive medal rate, more than doubling the previous best result. 
Moreover, ML-Master achieved these remarkable results within just 12 hours, which is half the time typically allocated in previous studies. 
These results not only highlight the potential of ML-Master in solving complex AI development problems, but also emphasize its capability to accelerate the path toward AI4AI.

The integration of exploration and reasoning within ML-Master offers a robust framework for AI systems that can autonomously evolve, learn, and adapt to increasingly complex challenges. As such, this work represents an important step in advancing AI4AI technologies. In future work, we aim to further refine the scalability and adaptability of ML-Master, particularly in dynamic and multi-agent environments, to continue pushing the boundaries of AI agent autonomy and generalization.

\clearpage
{\small
\bibliographystyle{unsrt}
\bibliography{ref}
} 

\appendix

\end{document}